# From Human to Robot Interactions: A Circular Approach towards Trustworthy Social Robots

Anna L. Lange[1,2], Murat Kirtay [3], Verena V. Hafner[1,2]

*Abstract*— Human trust research uncovered important catalysts for trust building between interaction partners such as appearance or cognitive factors. The introduction of robots into social interactions calls for a reevaluation of these findings and also brings new challenges and opportunities. In this paper, we suggest approaching trust research in a circular way by drawing from human trust findings, validating them and conceptualizing them for robots, and finally using the precise manipulability of robots to explore previously less-explored areas of trust formation to generate new hypotheses for trust building between agents.

## I. INTRODUCTION

During human-human interactions (HHI), individuals typically develop an intuition regarding the trustworthiness of their interaction partners. This intuition is often based on various indicators, including voice [1], [2], physical appearance [3], [4], body language [5]–[7], and emotional cues [8]. The initial impression formed is subjective and influenced by personal factors, such as the perceived similarity between oneself and the partner [9], [10]. As interactions become more complex or frequent, additional factors come into play, such as consistency in behavior [11], [12], truthfulness of statements [13], and mutual success during the interactions [14], [15]. Drawing from the established conceptualization of trust by Lewis and Weigert, trust can be understood as comprising three distinct dimensions: cognitive, affective, and behavioral [16]. These dimensions interact and influence the formation of trust.

In the context of advancing social robots, there is continuous research on how our understanding of trust from human psychology can be applied to human-robot interactions (HRI). However, this field faces several challenges, one being the three directions of trust in interactions involving robots and humans. Firstly, humans should be able to develop trust towards robots (human trust in robots) to allow for collaborative interactions. Secondly, robots should be able to decide whether or not their interaction partner is trustworthy (robot trust in humans) to increase their task effectiveness.

Finally, including the above two, bidirectional trust can develop (human-robot reciprocal trust).

The primary and extensively studied question revolves around how humans can develop trust in robots [17], [18]. Studies investigating this direction of trust often aim to advance human-robot teams for collaborative task completion, aiming to transition the robot's role from a mere tool to that of a teammate [17]. While individual factors such as a robot's appearance and behavior contribute to trust development, other contextual and social factors also play important roles [19], [20]. For instance, the human partner's gender might influence the level of trust placed in robots across different situations, as suggested by Cocchella and colleagues' study involving interactions between children and the iCub robot [18]. While the participants in this HRI study were children, this phenomenon is known from studies on human trust with adult participants (e.g. [21]). Furthermore, ensuring safety is a vital factor for fostering trust in robots and thus remains a central focus in research pertaining to trustworthy robots (e.g., see [22]). This concern encompasses not only physical safety but also the prevention of unrecognizable false outputs, often addressed through enhancing the transparency of artificial systems [23].

Examples of trust frameworks for robots, enabling them to determine whether to trust human or robot partners, have recently gained attention among researchers. The effectiveness of these frameworks carries implications for human trust in robots. Interaction with a partner that has the ability to trust often leads to reciprocal behavior and, consequently, increased trust in the partner [16]. Moreover, the facilitation of accurate trustworthiness assessments by a robot or agent could serve as a preventive measure against adverse manifestations like the Microsoft Tay incident [24]. Particularly in the context of joint tasks, the ability to select a reliable partner significantly influences the task outcome [25]. However, only few researchers are currently devoted to developing mechanisms for robots to cultivate trust in their partners. One of the first trust models for artificial systems employed a probabilistic framework based on Bayesian Networks and demonstrated results comparable to real data [26]. In the context of collaborative tasks, Vinanzi and colleagues proposed an artificial trust model incorporating an artificial episodic memory system and a Theory of Mind module [25]. More recently, a robotic trust module has emerged, utilizing an artificial neural network and a reinforcement learning algorithm to model trust recognition based on incurred cognitive load on the robot [27]. Their framework was also successfully employed during robot-robot interactions (RRI)

*This work was funded by the Deutsche Forschungsgemeinschaft (DFG, German Research Foundation) under Germany's Excellence Strategy - EXC 2002/1 "Science of Intelligence" - project number 390523135.

[1] Anna L. Lange, and Verena V. Hafner are with Adaptive Systems Group, Department of Computer Science, Humboldt-Universität zu Berlin, Berlin, Germany. hafner@informatik.hu-berlin.de
[2] Anna L. Lange, and Verena V. Hafner Science of Intelligence, Research Cluster of Excellence, Berlin, Germany. anna.lange@bccn-berlin.de
[3]Murat Kirtay is with the department of Cognitive Science and Artificial Intelligence, Tilburg University, Tilburg, Netherlands. m.kirtay@tilburguniversity.edu

[28]. Human trust ratings can be evaluated through questionnaires; however, a more quantitative approach is necessary for constructing trust in robots. The aforementioned study employs the cost of perceptual processing as an indicator of cognitive load, drawing inspiration from sources such as [29]. While the nature of trust is multifaceted and it is therefore important to acknowledge the challenge of attaining a comprehensive definition, our approach to defining trust is grounded in the 'balanced' trust definition introduced by [30], [31] as it has been shown to be applicable in robots as well as humans [25]. Following this definition, x represents a cognitive agent characterized by explicit goals (e.g., to diminish cognitive load), while y signifies an agent entrusted by x to assume certain delegated actions. In an interactive context, x places trust in y when the actions carried out by y prove to be beneficial for x. Here both x and y can be human or robotic agents.

The aforementioned models were developed mostly along methods from computer science combined with verification methods from engineering. To the best of our knowledge, there are no results from a psychological perspective on how a human observer would assess the trustworthiness of a robot equipped with one of the proposed frameworks. This brings us to the third and final type of trust in HRI: human-robot reciprocal trust, which has been discussed by Zonca and Sciutti [32]. We put forward that this direction of HRI trust research, which has seen relatively little interest, has great potential not only for improving social robots but also for new insights into trust development in general. We suggest adopting a circular approach in investigating trust in HRI. This approach encompasses multiple stages, beginning with an examination of human trust-related findings. These findings are then rigorously tested within the context of HRI, employing robots as the experimental subjects. Subsequently, the manipulability of robots is leveraged to conduct reciprocal trust studies, thereby paving the way for the identification of new hypotheses that can be tested in future human trust studies. Figure 1 provides a visual representation of this proposed circular approach.

## II. Implementation

This workflow is employed in HRI trust studies, yet unexplored gaps with potential insight remain. Cultivating trust toward social robots should start with insights from human interactions. Initially, robots are unfamiliar entities, and as Lewis and Weigert noted in their 1985 paper, "When faced with the totally unknown, we can gamble but we cannot trust" (see [16], p. 970). Consequently, for a robot to gain trust, it must transition from being an unknown entity to a known one. Incorporating humanlike features or behaviors into robots, drawing inspiration from human partners, and assessing their impact on trust ratings facilitate this process [33]. This principle can be expanded beyond external features to encompass broader trust concepts. In a study conducted by Samson and Kostyszyn, participants' trust was evaluated during an economic game interaction with one or two secondary tasks [34]. Their findings concluded that the cognitive load experienced during a task influences the level of trust individuals place in their interaction partner. These human to human trust findings were replicated in the field of HRI, where cognitive load during a task affected the level of trust assigned to a robot by human participants [29], [35]. Thereby we are entering the first step in the loop for cyclical trust studies: by observing behavior from human to human trust formation, similar scenarios were implemented with a robot partner and the human to robot trust response was assessed. Building on these findings, and entering the next step of the loop a robotic trust framework was developed, incorporating an associative memory module that provides a cognitive load value during a task and enabling a robot to form trust in a human [27]. Subsequently, Taliaronak and colleagues expanded this concept by including a social dimension, enabling a robot to assess the trustworthiness of its human partner based on their gestures, combined with the perceived cognitive load, to determine a trust rating [36]. The researchers followed an engineering workflow, encompassing framework design, implementation, and verification. However, one crucial aspect that has not been considered following the development of this model is the evaluation of how a human observer's or interaction partner's trust ratings are affected by the robotic trust framework. Does the inclusion of an artificial trust module contribute to increased reciprocal trust building between the human and the robot? This would be a critical next step in completing the full circle from human to robot and back. Lastly, research outcomes from studies that incorporate an adaptive framework, enabling deliberate manipulation of parameters, can contribute valuable insights into our fundamental comprehension of the human trust formation process. Investigating long-term trust development and changes is particularly challenging to conduct under controlled conditions with human participants. By having functional and evaluated frameworks that can be implemented in robots, we gain the ability to conduct experiments with one controllable partner (HRI) or two controlled partners (RRI). This advancement paves the way for exploring previously unanswered questions such as, 'How does trust evolve and transform between individuals over time?'. Therefore completing each step in the cycle allows us to build biologically inspired and verified trust frameworks for robots and simultaneously enter new realms for understanding the human biology.

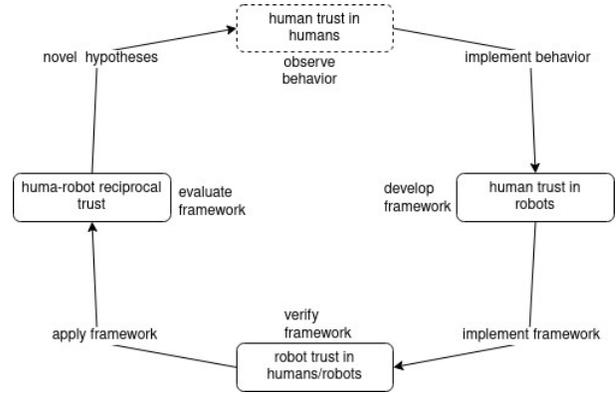

Fig. 1: The circular flow for human-robot trust studies involves the following concise steps: 1. Derive trustworthy behavior insights from human studies. 2. Implement the identified behavior on a robot and test its effects. 3. Develop robotic trust frameworks based on the obtained insights, providing structures for different models. 4. Test these frameworks in reciprocal trust studies to evaluate their impact on human interaction partners, where artificial agents can also assume the role of the human. 5. Incorporate the insights gained from these studies into new human trust studies, continuing the circular process.

## III. Discussion

HRI researchers from diverse backgrounds, including psychology, cognitive sciences, computer science, engineering, neuroscience, and ethical and legal studies, adopt an interdisciplinary approach, creating a broad knowledge base for understanding trust in HHI, HRI, and RRI [37]. Nevertheless, the current literature reveals deficiencies in certain domains, including the translation of human trust research findings into robot implementation and the investigation of the impacts of established implementations on human interaction partners. With this paper, we aim to inspire future research collaborations between disciplines (e.g. psychologists and computer scientists), utilizing the wealth of available knowledge to investigate trust in a cyclical manner, encompassing interactions from humans to robots and back.